\documentclass[10pt,twocolumn,letterpaper]{article}
\usepackage{cvpr}              
\usepackage{epsfig}
\usepackage{graphicx}
\usepackage{amsmath}
\usepackage{amssymb}
\usepackage{multirow}
\usepackage{enumitem}
\usepackage{textcomp,booktabs}
\usepackage{algorithm}
\usepackage{algpseudocode}
\usepackage[dvipsnames]{xcolor}
\usepackage{colortbl}
\newcommand{\myparagraph}[1]{\vspace{1pt}\noindent{\bf{#1}}}
\usepackage{diagbox}
\usepackage[pagebackref,breaklinks,colorlinks]{hyperref}
\usepackage[capitalize]{cleveref}
\crefname{section}{Sec.}{Secs.}
\Crefname{section}{Section}{Sections}
\Crefname{table}{Table}{Tables}
\crefname{table}{Tab.}{Tabs.}

\def\confName{CVPR}
\def\confYear{2022}

\begin{document}

\title{Probabilistic Compositional Embeddings for Multimodal Image Retrieval}

\author{
Andrei Neculai\textsuperscript{1}, \hspace{3pt}
Yanbei Chen\textsuperscript{1}, \hspace{3pt}
Zeynep Akata\textsuperscript{1,2,3} \\
{\small
\textsuperscript{1}University of T{\"u}bingen \hspace{2pt}
\textsuperscript{2}MPI for Informatics \hspace{2pt} 
\textsuperscript{3}MPI for Intelligent Systems} \\
{\small neculai.andrei0@gmail.com, \{yanbei.chen, zeynep.akata\}@uni-tuebingen.de}
}

\maketitle

\begin{abstract} 
Existing works in image retrieval often consider retrieving images with one or two query inputs, which do not generalize to multiple queries. In this work, we investigate a more challenging scenario for composing multiple multimodal queries in image retrieval. Given an arbitrary number of query images and (or) texts, our goal is to retrieve target images containing the semantic concepts specified in multiple multimodal queries. To learn an informative embedding that can flexibly encode the semantics of various queries, we propose a novel multimodal probabilistic composer (MPC). Specifically, we model input images and texts as probabilistic embeddings, which can be further composed by a probabilistic composition rule to facilitate image retrieval with multiple multimodal queries. We propose a new benchmark based on the MS-COCO dataset and evaluate our model on various setups that compose multiple images and (or) text queries for multimodal image retrieval. Without bells and whistles, we show that our probabilistic model formulation significantly outperforms existing related methods on multimodal image retrieval while generalizing well to query with different amounts of inputs given in arbitrary visual and (or) textual modalities. Code is here: 
\url{https://github.com/andreineculai/MPC}. 
\end{abstract}

\vspace{-1em}
\section{Introduction}
\begin{figure}[!ht]
\begin{center}
\includegraphics[width=\linewidth]{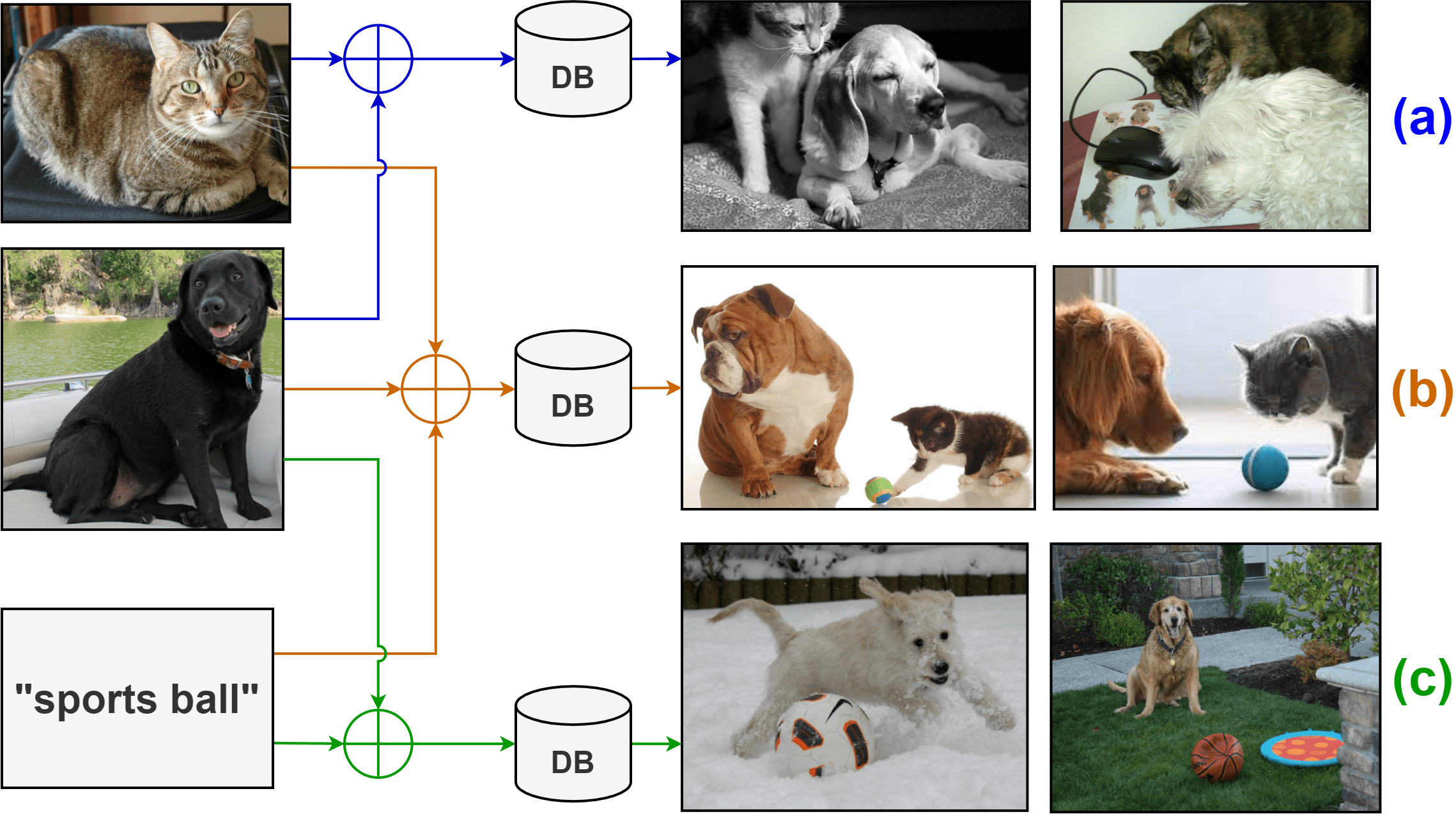}
\end{center}
\vskip -1em
\caption{
We consider a new compositional learning paradigm for multimodal image retrieval. Given an {\em arbitrary} number of multimodal queries (e.g. images and texts), the model is trained to learn a feature embedding for retrieving images that contain the composite set of semantic concepts specified in the queries, where the input could be given in {\em arbitrary} combinations such as (a) image + image, (b) image + image + text, (c) image + text, etc.
}
\label{fig:teaser}
\vskip -1em
\end{figure}

Image retrieval aims to learn an embedding that encodes the semantics of an input query for retrieving the most relevant target images in the database. Existing works in image retrieval often consider taking one single image \cite{gordo2016deep} or a text sentence \cite{frome2013devise} as query, which limit their applicability in coping with multiple queries given in arbitrary modalities. In this work, we investigate a more challenging scenario in multimodal image retrieval, where the queries may consist of an arbitrary number of queries in arbitrary visual and textual modalities (e.g. images and texts). 

As Figure \ref{fig:teaser} shows, given an arbitrary number of image and (or) text queries, our goal is to retrieve the images that contain all the semantic concepts specified in the queries. Inspired by the recent advances in compositional learning for visual recognition \cite{purushwalkam2019task,naeem2021learning,mancini2021open}, we tackle this problem by learning a compositional embedding to flexibly encapsulates the multiple semantic concepts specified in the multimodal queries, and to be used for retrieving the more relevant images. For instance, when giving two query images capturing ``{\em cat}'' and ``{\em dog}'', and one text query stating ``{\em sports ball}'', we aim to learn a compositional embedding that represents all three queries to retrieve images that contain ``{\em cat}'', ``{\em dog}'' and ``{\em sports ball}'' (see Figure \ref{fig:teaser} (b)). 

A line of recent works \cite{Vo_2019_CVPR,chen2020image,hosseinzadeh2020composed,chen2020learning,wu2021fashion} also explore compositional learning upon image and text. However, all these existing works consider evaluating and modelling upon only {\em two} queries given in a {fixed} combination of image and text. Moreover, these works mainly consider using text to specify the modifications in input image for retrieving similar images, which does not apply to the case when the retrieved images are expected to represent a composite set of multiple semantic concepts. 
Therefore, to further investigate image retrieval using an {\em arbitrary} number of queries (e.g. more than {two} queries) given in {\em arbitrary} visual and (or) textual modalities, we establish a new evaluation benchmark for multimodal image retrieval. 
Our new benchmark is built with the MS-COCO dataset \cite{lin2015microsoft} and offers a more advanced and challenging testbed to facilitate research on image retrieval. 
We propose this new benchmark with a comprehensive evaluation of multiple state-of-the-art multimodal methods and propose a novel Multimodal Probabilistic Composer ({\bf MPC}) that learns an informative probabilistic compositional embedding to flexibly encode an arbitrary number of queries. Specifically, rather than fusing embeddings from different modalities through a fixed set of learnable parameters, we propose to parameterize each input as a probabilistic embedding following a multivariate Gaussian, and compose embeddings by deriving a composite multivariate Gaussian based on a probabilistic composition rule. 

Our model formulation offers two unique properties to learn a compositional embedding for multimodal image retrieval. First, our probabilistic composer allows to compose embeddings of a flexible amount of queries in arbitrary modalities. Second, its probabilistic nature allows to encode semantics as well as ambiguities of a given input, thus well capturing the polysemantic information in text queries, e.g. a text query ``{\em dog}'' may refer to a variety of dog breeds that differ visually. These properties well faciliate better performance in multimodal image retrieval. 

Our {\bf contribution} is three-fold: 
\vspace{-0.3em}
\noindent
\begin{itemize}[noitemsep,topsep=5pt,itemsep=2pt,leftmargin=12pt]
\item We establish a new multimodal image retrieval benchmark using the MS-COCO dataset to investigate image retrieval using an arbitrary number of queries in arbitrary modalities. We evaluate a variety of settings including (1) using different combinations of input modalities, and (2) using various number of queries. 
\item We propose a Multimodal Probabilistic Composer (MPC), which features a new probabilistic rule to compose probabilistic emebddings and a new probabilistic similarity metric to compare probabilistic embeddings, which together lead to its superior model performance in composing multimodal queries for image retrieval. 
\item We show that our model outperforms existing multimodal fusion methods significantly for mulitmodal image retrieval. To further analyze our model design rationale, we also conduct an indepth experimental analysis.
\end{itemize}

\section{Related Work}

\myparagraph{Multimodal image retrieval} refers to image retrieval with multimodal queries. In contrast to standard image retrieval that takes in one image as query, multimodal image retrieval digests multiple inputs in different modalities such as images, spatial layout~\cite{mai2017spatial}, sketch~\cite{yu2016sketch,yelamarthi2018zero,ghosh2019interactive}, texts in the attribute formats \cite{parikh2011relative,kovashka2012whittlesearch,yu2019thinking,zhao2017memory,han2017automatic,ak2018learning} or natural language sentences \cite{guo2018dialog,guo2019fashion,chen2020image,hosseinzadeh2020composed,chen2020learning,wu2021fashion}. Unlike these existing works that consider the other non-visual modalities as feedback to guide image retrieval, we focus on learning the compositions of multiple semantic concepts specified in the query.

\myparagraph{Compositional learning} provides an important functionality in an artificial intelligent machine~\cite{lake2015human,lake2017building}. The goal of compositionality is to learn representations or data distributions that encapsulate multiple semantic concepts
~\cite{misra2017red,nagarajan2018attributes,tokmakov2019learning,purushwalkam2019task,wei2019adversarial,naeem2021learning,tan2019text2scene,Vo_2019_CVPR,materzynska2020something,peyre2019detecting}. One line of works explore compositionality in zero-shot recognition \cite{purushwalkam2019task,naeem2021learning,mancini2021open}, where the model is trained to compose states and objects (e.g. ``{\em ripe}'' and ``{\em apple}'') and used is to recognize their compositions. 
Another line of works introduce compositionality for image generation, which aim to synthesize new images that contain composite visual primitives \cite{park2021benchmark,tan2019text2scene,azadi2020compositional,tseng2020retrievegan,niemeyer2021giraffe} or composite context and semantic content \cite{xue2012understanding,tsai2017deep,lin2018st,chen2019instance}. 
More recently, compositionality is explored in various tasks such as image retrieval \cite{Vo_2019_CVPR,chen2020image,hosseinzadeh2020composed}, 
video recognition \cite{materzynska2020something,chen2021distilling}, and visual relationship detection \cite{peyre2019detecting,hou2020visual}. 
In this work, we study a new compositional learning paradigm for multimodal image retrieval. Unlike the relevant works in image retrieval that consider composing image and text for modifying image content \cite{Vo_2019_CVPR,chen2020image,hosseinzadeh2020composed} , we establish a new benchmark to investigate compositions upon an {\em arbitrary} number of semantics concepts given in {\em arbitrary} modalities. Moreover, we propose a novel probabilistic model formulation to derive probabilistic compositional embeddings that jointly capture semantics and ambiguities in input data to learn more informative compositional embeddings. 

\myparagraph{Multimodal fusion} is widely adopted as an effective learning scheme to fuse complementary and (or) compositional information from multimodal input data for building a more powerful model. A large body of works exploit multimodal fusion to enhance video understanding, such as utilizing audio and visual data  \cite{wang2020makes,liu2019use,chen2021distilling,jaegle2021perceiver,nagrani2021attention}, or using text and visual data  \cite{sun2019videobert,gabeur2020multi} to learn video representations. A branch of recent works leverage multimodal fusion to fuse RGB images with depth or LiDAR input to enhance representation learning in autonomous driving \cite{xiao2020multimodal,prakash2021multi}. Another volume of works use multimodal fusion to integrate images and texts for solving multimodal image retrieval \cite{Vo_2019_CVPR,chen2020image,hosseinzadeh2020composed,chen2020learning,wu2021fashion} or visual question answering (VQA) \cite{antol2015vqa,teney2018tips}. Unlike existing works that fuse a fixed combinations of modalities, we explore multimodal fusion upon an arbitrary composite set of semantic concepts specified in different data modalities. 

\begin{figure*}[!ht]
\begin{center}
\includegraphics[width=.95\textwidth]{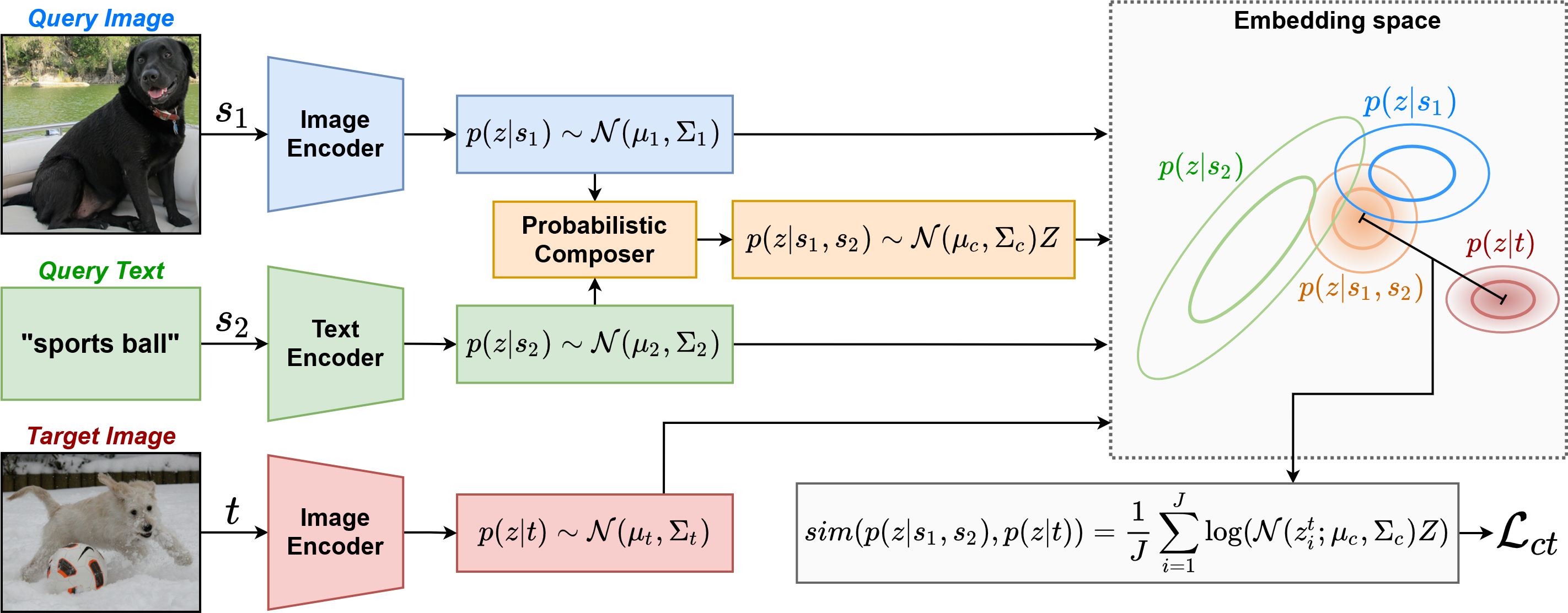}
\end{center}
\vskip -1.5em
\caption{
Model overview. We show how Multimodal Probabilistic Composer (MPC) learns from {\em two} queries for {\em visual simplicity}. Here, given the query image $s_1$ and text $s_2$, MPC first learns two modality-specific probabilistic embeddings (Section \ref{sec:modality_specific_embedders}), and then composes the two embeddings through a probabilistic composer to obtain a probabilistic compositional embedding (Section \ref{sec:probabilistic_compositions}) which is further aligned with the probabilistic embedding of the target image $t$ by minimizing a probabilistic distance metric (Section \ref{sec:model_optimization}). Note: MPC can digest {\em more than two} queries by applying the probabilistic composer on a set of probabilistic embeddings (detailed in Section \ref{sec:probabilistic_compositions}). 
}
\label{fig:overview}
\vskip -0.5em
\end{figure*}

\section{Multimodal Probabilistic Composer}

Given a composite set of $k$ input samples -- where each input specifies a semantic concept (e.g. ``{\em dog}'', ``{\em sport ball}''), our goal is to learn a compositional embedding for retrieving the corresponding target images that contain the set of specified semantic concepts. Each input can be given in visual or textual modality, and passed through modality-specific encoder to obtain its embedding. Accordingly, the composition of different embeddings should be modality-agnostic, which means our model by default is trained to combine an arbitrary set of samples in arbitrary modalities.

Figure \ref{fig:overview} gives an overview of our proposed model:  Multimodal Probabilistic Composer (MPC). For simplicity, we visualize the case where MPC takes in two queries (a ``{\em dog}'' image and a ``{\em sport ball}'' text snippet) to retrieve a target image that contains ``{\em dog}'' and ``{\em sport ball}''. The image and text are passed through modality-specific encoders to learn probabilistic embeddings (Section \ref{sec:modality_specific_embedders}) and then composed to derive a compositional embedding (Section \ref{sec:probabilistic_compositions}). MPC is optimized to align the compositional embedding and the target image embedding by minimizing a probabilistic contrastive loss (Section \ref{sec:model_optimization}). While Figure \ref{fig:overview} shows the scenario with two inputs, MPC is designed with the flexibility to process an arbitrary set of samples in arbitrary modalities thanks to its probabilistic nature.

\subsection{Modality-Specific Probabilistic Embeddings}
\label{sec:modality_specific_embedders}

In this section, we first describe how different modalities (i.e. image and text) are modeled and then detail how probabilistic embeddings are learned to represent each input. 

\myparagraph{Image encoder.} 
We use a ResNet \cite{he2016deep} backbone $f_{\text{ResNet}}$, with an additional linear projection layer $f_{img}$, as our image encoder to learn the image embeddings. Given an input image (referred as $s_1$), we pass it through $f_{\text{ResNet}}$ to obtain the feature map $\phi_{img}$ and compute its feature encoding as: 
$z_{img} = f_{img}(\phi_{img})$, where $z_{img} \in \mathbb{R}^D$.

\myparagraph{Text encoder.} 
To encode text information, we use GloVe word embeddings \cite{pennington2014glove} to encode each word (denoted as $f_{GloVe}$) and train a bidirectional GRU \cite{cho2014properties} (denoted as $f_{txt}$) to learn the text embeddings. Given a text snippet (referred as $s_2$), we obtained the word embeddings: $\phi_{txt} = f_{GloVe}(s_2)$. We pass $\phi_{txt}$ through the GRU to obtain its feature encoding: $z_{txt} = f_{txt}(\phi_{txt})$, where $z_{txt} \in \mathbb{R}^D$.

\myparagraph{Probabilistic embeddings.} Our model is motivated with the flexibility to take a composite set of $k$ queries in arbitrary modalities. With this design rationale in mind, we propose to model each embedding as a multivariate Gaussian probability density function (PDF), such that the compositions of different embeddings can be achieved by composing different Gaussian PDFs through a parametric probabilistic rule, i.e. the product of $k$ Gaussian PDFs \cite{bromiley2003products}. Below, we detail how each embedding is modeled as a multivariate Gaussian, similar to recent probabilistic embeddings works \cite{oh2018modeling,chun2021pcme}, and present how we derive compostional embeddings through our probabilistic composer in Section \ref{sec:probabilistic_compositions}. 

Given any embedding and feature map (denoted as $z_m$ and $\phi_m$) from either the image encoder or the text encoder, we model it as a Gaussian with mean $\mu_m$ and diagonal variance matrix $\Sigma_m$. For computational reasons, we work with only the diagonal of the covariance matrix $\Sigma_m$, referred as $\sigma_m^2$. As seen in \cite{chun2021pcme}, $\mu_m$ and $\sigma_m^2$ are computed as follows: 
\begin{equation}
\label{eq:prob_emb}
\begin{aligned}
&\mu_m = \text{LN}((z_m + s(\text{fc}(\text{attn}(\phi_m))))\\
&\log (\sigma_m^2) = z_m + \text{fc}(\text{attn}(\phi_m))
\end{aligned}
\end{equation}
where LN denotes a LayerNorm \cite{ba2016layer} operation, fc denotes a linear layer, $s(\cdot)$ is a sigmoid activation function and attn denotes a self-attention module \cite{lin2017structured}. The model outputs the logarithm of the variance instead of the actual variance for numerical stability. 
In essence, the Gaussian PDF (Eq. \ref{eq:prob_emb}) offers a more informative and expressive estimation about the embedding,  capturing not only the semantics of an input but also its ambiguities. By design, imposing a probabilistic distribution on a feature embedding has the following merits. First, it is essential in practice to encode both semantics and ambiguities of an input -- e.g. encoding a query of “{\em cat}” needs to represent its semantic content as a cat but also need to represent its ambiguities to reflect 
different possible variations as not all cats visually look the same. Second, composing a set of $k$ probabilistic embeddings can be easily derived by composing their Gaussian PDFs through a parametric probabilistic formulation, as detailed next.  

\subsection{Probabilistic Composer}
\label{sec:probabilistic_compositions}

\myparagraph{Probabilistic composer.} Let $S{=}\{s_1, \dots s_k\}$ be the set of $k$ samples in the query. The goal of the composition is to find a probability distribution $p(z|S)$ that unifies all the individual probability distributions $\{p(z|s_i) \sim \mathcal{N}(\mu_i, \Sigma_i)\}_{1,k}$. As aforementioned, composing a set of $k$ probabilistic embeddings can be achieved based on a parametric probabilistic rule which derives the product of $k$ Gaussian PDFs \cite{bromiley2003products}. We refer this process as our {probabilistic composer} and explain the case where $k=2$ (i.e. two queries) below. Formally, the product of two Gaussian PDFs can be written as:
\begin{equation}
\label{eq:gaussian_multiplication}
\begin{aligned}
\mathcal{N}(z;\mu_1, \Sigma_1&)\mathcal{N}(z;\mu_2, \Sigma_2) = \mathcal{N}(z;\mu_c, \Sigma_c)Z
\end{aligned}
\end{equation}
where $\mathcal{N}(z;\mu_1, \Sigma_1), \mathcal{N}(z;\mu_2, \Sigma_2)$ are the two Gaussian PDFs for the set of two queries $s_1, s_2$. $\mathcal{N}(z;\mu_c, \Sigma_c)$ is the new composite Gaussian with mean $\mu_c$ and variance $\Sigma_c$, and $Z$ is a normalization constant, defined as:
\begin{equation}
\label{eq:gaussian_multiplication2}
\begin{aligned}
\Sigma_c &= (\Sigma_1^{-1} + \Sigma_2^{-1})^{-1}\\
\mu_c &= \Sigma_c(\Sigma_1^{-1}\mu_1 + \Sigma_2^{-1}\mu_2)\\
Z &= \mathcal{N}(\mu_1;\mu_2, \Sigma_1 + \Sigma_2)
\end{aligned}
\end{equation}
While Eq. \ref{eq:gaussian_multiplication}, Eq. \ref{eq:gaussian_multiplication2} show the special case of $k=2$, it can be easily generalized to the case where $k > 2$ by deriving the product of $k$ Gaussian PDFs sequentially -- i.e. multiplying the $k_{th}$ Gaussian PDF with the product of all the $(k-1)$ Gaussian PDFs, which results in a composite new Gaussian with a normalization constant $Z$ written as follows.
\begin{equation}
\label{eq:resulting_z}
\begin{aligned}
Z &= \prod_{i=1}^{k-1} Z_{i,i+1}\\
\end{aligned}
\end{equation}
where $Z_{i,j}$ is the normalization constant of the product between the Gaussian PDFs of input $i$ and input $j$. Thanks to the inductive bias induced by our probabilistic composer, our model can generalize to compositions of a flexible set of queries even without training. In our experiments, we show that despite not containing any learned parameters, our probabilistic composer outperforms other composition methods that learn fusion layers to compose embeddings. 

\subsection{Model Optimization}
\label{sec:model_optimization}

Similar to standard objectives in non-probabilistic metric learning such as triplet loss and contrastive loss, our training objective is imposed to pull the distribution of the compositional embedding and the target image distribution closer, while pushing away the distributions of negative pairs. To achieve this aim, we first need to define a probabilistic similarity function between two probability distributions.

\myparagraph{Probabilistic similarity.} To quantify the similarity between two probabilistic distribution, Monte-Carlo estimation can be adopted \cite{chun2021pcme}, which draws a number of $J$ data points $\{z_i^x\}_{1,J}$, $\{z_i^y\}_{1,J}$ from two distributions $p(z|x), p(z|y)$, resulting in $J^2$ pairs of feature vectors that can be used to compute the similarity between two distributions as: 
\begin{equation}
\label{eq:similarity_function_pcme}
\begin{aligned}
sim(p(z|x), p(z|y)) = \frac{1}{J^2} \sum_{i=1}^{J}\sum_{j=1}^{J} \kappa(z_i^x, z_j^y) 
\end{aligned}
\end{equation}
where $\kappa(\cdot, \cdot)$ refers to a standard similarity metric between feature vectors, e.g. cosine similarity. 
Although Eq. \ref{eq:similarity_function_pcme} can measure the similarity between two probabilistic embeddings that follow Gaussian distributions, it could not be directly applied in our scenario given that our composite Gaussian is scaled by a constant $Z$ (Eq. \ref{eq:resulting_z}); while sampling data points from $\mathcal{N}(z;\mu_c, \Sigma_c)$ is equivalent to sampling from $\mathcal{N}(z;\mu_c, \Sigma_c)Z$. 

To align the probabilistic compositional embedding (Eq. \ref{eq:gaussian_multiplication}) -- modeled as a composite Gaussian $\mathcal{N}(z; \mu_c, \Sigma_c)Z$ -- and the target image embedding -- modeled as a multivariate Gaussian $\mathcal{N}(z; \mu_t, \Sigma_t)$), we propose a new probabilistic similarity function to measure the similarity between the composite Gaussian $p(z|S)$ and the target image distribution $p(z|t)$ using the following procedure. First, we sample $J$ data points $\{z_i^t\}_{1,J}$ from the target distribution $\mathcal{N}(\mu_t, \Sigma_t)$. Then, we assign these data points to the composite distribution $\mathcal{N}(z;\mu_c, \Sigma_c)Z$ to compute the probabilistic similarity scores. 
The similarity function is formally defined as the logarithm of these probabilistic scores: 
\begin{equation}
\label{eq:similarity_function}
\begin{aligned}
sim(p(z|S), p(z|t)) = \frac{1}{J} \sum_{i=1}^{J} \log (\mathcal{N}(z_i^t; \mu_c, \Sigma_c)Z)\\
\end{aligned}
\end{equation}
where we compute the overall similarity with logarithm to ensure numerical stability given that $Z$ may contain small numerical values. In essence, Eq. \ref{eq:similarity_function} provides a tractable probabilistic formulation to measure the similarity between two probabilistic embeddings that follow the Gaussian distributions $\mathcal{N}(z; \mu_c, \Sigma_c)Z$ and $\mathcal{N}(z; \mu_t, \Sigma_t)$. Eq. \ref{eq:similarity_function} is also computationally more efficient than Eq. \ref{eq:similarity_function_pcme}, given that their computational complexity is $O(J)$ vs $O(J^2)$. 

\myparagraph{Learning objective.} In similar spirit as the cross entropy loss and contrastive loss \cite{chen2020simple}, we define our loss function as: 
\begin{equation}
\label{eq:loss_function}
\begin{aligned}
\mathcal{L}_{ct}{=}\frac{1}{B} \sum_{i=1}^{B}{-}\log\frac{\text{exp}(sim(p(z|S_i), p(z|t_i)))}{\sum_{j=1}^{B}\text{exp}(sim(p(z|S_i), p(z|t_j)))} \\
\end{aligned}
\end{equation}
where $B$ denotes the batch size. $sim(p(z|S_i), p(z|t_i))$ is the similarity between probabilistic distributions of two positive pairs. The loss $\mathcal{L}_{ct}$ is computed across all positive pairs.
To ensure the training stability and prevent the learned variance $\sigma_m^2$ (Eq. \ref{eq:prob_emb}) from collapsing to zero or exploding to very high values, we add a $\ell_2$ regularization term on the logarithm of the variance, as defined below: 
\begin{equation}
\label{eq:l2_loss_function}
\begin{aligned}
\mathcal{L}_{\ell_2}=\frac{1}{B|S|} \sum_{i=1}^{B}\sum_{j=1}^{S} (\log (\sigma_{i,j}^2))^2 \\
\end{aligned}
\end{equation}
where $|S|$ is the number of queries being composed. 
$\sigma_{i,j}^2$ is the variance of the input $j$ in the $i_{th}$ pair of queries in the batch. 
The final loss is $\mathcal{L}=\mathcal{L}_{ct}+\lambda_{\ell_2}\mathcal{L}_{\ell_2}$.

\myparagraph{Model deployment.} 
At test time, we compute the compositional probabilistic embeddings of the set of queries by first computing the probabilistic embeddings (Eq. \ref{eq:prob_emb}) and then composing the embeddings through our probabilistic composer (Eq. \ref{eq:gaussian_multiplication}, Eq. \ref{eq:gaussian_multiplication2}). For image retrieval, we compute the probabilistic embeddings for all the images in the database, and match the composite queries with all images based on the cosine similarity between the mean $\mu_c$ of a compositional probabilistic embeddings and the mean $\mu_t$ of an image embedding. The images with the top-$k$ similarity scores in the database are the top-$k$ retrieved items. 

\section{Experiments}

\subsection{Constructing Benchmark}
\label{sec:setup}

\myparagraph{Dataset.}
To explore compositional learning upon a flexible combination of queries, we establish a new benchmark based on the MS-COCO dataset \cite{lin2015microsoft}, which contains bounding box annotations of 80 unique object categories. 
We utilize these data annotations to construct our multimodal image retrieval benchmark. As each image in MS-COCO often contains a set of objects labeled with bounding boxes, we can easily obtain {\em full images} that contain a composite set of objects, as well as {\em cropped images} that contain a single object by applying the bounding boxes. Concretely, we use the {\em cropped images} with one object (e.g. {\em dog}, {\em cat} and {\em truck}) as the query images, and use the category label (e.g. {\em sport balls}, {\em carrot} and {\em stop sign}) as the query texts. It is worth noting that we consider short texts in this work; however, our model is able to handle free-form texts, given that the text encoder can encode both words and sentences. 
The query images and (or) query texts can be composed by composition methods to retrieve the target images that contain the composite set of specified semantic concepts. 
More details about the benchmark are given in the supplementary. 

\myparagraph{Benchmark setup.} To build a comprehensive benchmark, we consider various evaluation setups, as detailed next. 
\vspace{-0.5em}
\begin{itemize}[noitemsep,topsep=5pt,itemsep=1pt,leftmargin=10pt]
\item composing different input modalities: images only, texts only, and multimodal query with images and texts.
\item seen compositions vs unseen compositions: the former considers that the composite concepts encountered at test time are seen during training, while the latter considers the compositions are new and unseen at test time.
\item varying number of queries: 2, 3, 4 queries. 
\end{itemize}
\vspace{-0.3em}
We construct a training/validation/test data split with around 78,000/20,000/20,000 image samples using the MS-COCO dataset. 
For the seen compositions, we select 1000 category tuples for training and validation and we use the same tuples for testing. In the unseen compositions scenario we pick 100 pairs for training and validation and 500 different pairs for testing (the training and testing pairs use the same categories). Due to the high number of image samples for each category, the actual number of possible compositions of images and (or) texts is extremely high. Thus, the same compositions of queries are rarely repeated during training, offering rich diversity in the constructed query-target pairs for training. 
Due to space limit, we give data statistics and evaluate on other datasets in supplementary. 

\myparagraph{Evaluation metrics.} 
We use two metrics to evaluate image retrieval. First, we use recall@K (R@K), which measures the percentage of at least one correct retrieved images occurred in the top K retrieved items. Second, we use R-Precision (R-P) \cite{musgrave2020metric}, which takes into account that there are multiple correct retrieved images for each query, i.e.  $\text{R-P}{=}\frac{1}{N} \sum_{i=1}^{N}\frac{r}{R}$, where $N$ is the number of test queries; for each test query, $R$ is the number of correct target images in the gallery, and $r$ is the number of correct retrievals in the top-R retrieved items. 
$\text{R-P}$ has a high score only if the model ranks all the correct items before the incorrect ones.

\begin{table*}[!t]
	\small
	\centering
	\setlength{\tabcolsep}{10pt}
	\begin{tabular}{l|cc|c|cc|c|cc|c}
		\hline
		\multirow{2}{10pt}{Method} 
		& \multicolumn{3}{c|}{\texttt{images only}} 
		& \multicolumn{3}{c|}{\texttt{multimodal}} 
		& \multicolumn{3}{c}{\texttt{texts only}} \\ \cline{2-10}
		& R@5 & R@10 & R\textunderscore P & R@5 & R@10 & R\textunderscore P & R@5 & R@10 & R\textunderscore P  \\ 
		\hline
		 Relationship &
		 0.36 & 0.72 & 0.11 &
		 0.47 & 0.77 & 0.13 &
		 0.61 & 1.31 & 0.14 \\
		 FiLM  &
		 0.40 & 0.92 & 0.15 &
		 0.51 & 1.06 & 0.18 &
		 0.37 & 1.06 & 0.15\\
		 MRN & 
		 19.81 & 28.17 & 4.52 &
		 24.59 & 34.83 & 5.46 &
		 33.20 & 42.48 & 6.73 \\		 
		 TIRG & 
		 21.98 & 31.58 & 5.80 &
		 18.94 & 26.85 & 4.87 &
		 34.79 & 51.43 & 9.09 \\
		 PCME + addition & 
		 15.19 & 21.86 & 3.66 &
		 20.02 & 29.41 & 4.88 &
		 28.78 & 39.94 & 6.62\\
		 \hline \hline
         \bf MPC  & 
		 \bf 27.32 & 
		 \bf 36.52 & 
		 \bf 6.72 &
		 \bf 37.76 &
		 \bf 48.23 &
		 \bf 9.10 &
		 \bf 59.32 &
		 \bf 69.42 &
		 \bf 15.14 \\
		\hline
	\end{tabular}
	\vskip -1em
	\caption{
	Evaluation of composing {\bf \em two} queries for image retrieval on a {\bf \em seen} composition setup. 
	}
	\label{tab:2_categories_seen}
	\vskip -1em
\end{table*}

\begin{table*}[ht]
	\small
	\centering
	\setlength{\tabcolsep}{10pt}
	\begin{tabular}{l|cc|c|cc|c|cc|c}
		\hline
		\multirow{2}{10pt}{Method} 
		& \multicolumn{3}{c|}{\texttt{images only}} 
		& \multicolumn{3}{c|}{\texttt{multimodal}} 
		& \multicolumn{3}{c}{\texttt{texts only}} \\ \cline{2-10}
		& R@5 & R@10 & R\textunderscore P & R@5 & R@10 & R\textunderscore P & R@5 & R@10 & R\textunderscore P  \\ 
		\hline
		 Relationship &
		 0.40 & 0.49 & 0.13&
		 0.41 & 1.02 & 0.17 &
		 0.57 & 1.62 & 0.27\\
		 FiLM  &
         0.00 & 0.16 & 0.04&
		 0.01 & 0.12 & 0.05&
		 0.00 & 0.00 & 0.03 \\ 
		 MRN & 
		 \bf 9.89 & \bf 16.96 & \bf 2.21 &
		 17.63 & 26.97 & 3.73 &
		 25.16 & 36.48 & 5.14\\
		TIRG & 
		 6.95 & 11.97 & 1.62 &
		 4.77 & 8.93 & 1.30 &
		 2.91 & 5.01 & 0.76 \\
		 PCME + addition & 
		 4.61 & 8.00 & 1.09 &
		 7.37 & 12.78 & 1.68 &
		 10.44 & 16.26 & 2.08 \\
		 \hline \hline
         \bf MPC  & 
		  8.90 & 
		  15.13 & 
		  1.94 &
		 \bf 19.26 &
		 \bf 28.03 &
		 \bf 4.15 &
		 \bf 32.52 &
		 \bf 43.28 &
		 \bf 6.75\\
		\hline
	\end{tabular}
	\vskip -1em
	\caption{
	Evaluation of composing {\bf \em three} queries for image retrieval on a {\bf \em seen} composition setup. 
	}
	\label{tab:3_categories_seen}
	\vskip -1.5em
\end{table*}

\myparagraph{Implementation details.} 
We implement our model using PyTorch. For the image encoder, we use the ResNet-50 network \cite{deng2009imagenet} pre-trained on ImageNet. The 2048-D image embeddings are passed through a self-attention module with a hidden layer of dimension 1024 and a fc layer that maps it to a 512-D probabilistic embedding (Eq. \ref{eq:prob_emb}). 
For the text encoder, we use the GloVe 300-dimensional word embeddings pre-trained on the corpus with 1.9M vocabulary \cite{pennington2014glove}, and train a bidirectional GRU \cite{cho2014properties} with 256 features in its hidden state. The 256-D text embeddings are passed through a self-attention module with a hidden layer of dimension 150 and a fc layer that maps it to a 512-D probabilistic embedding (Eq. \ref{eq:prob_emb}). To compute the probabilistic similarity in Eq. \ref{eq:similarity_function}, we use $J=7$, same as \cite{chun2021pcme}. We set the weight $\lambda_{\ell_2}$ for the regularizer $\mathcal{L}_{\ell_2}$ to 0.001. For data augmentation, we use Cutout \cite{devries2017improved}, random resizing and random horizontal flips for the images and apply random word dropout with probability 0.1. 
For training, we use the Adam optimizer \cite{kingma2014adam} and our learning rate is 2e-4. We set the learning rate of ResNet layers before the GAP layer to 1/10 the general learning rate. We train our model for 1600 epochs and decrease the learning rate by a factor of 1/10 every 600 epochs. We use the same data augmentation and training scheme for different methods to ensure fair comparisons. 

\subsection{Comparing to the State-of-the-Art}
\label{sec:sota}

\myparagraph{Competitors.} 
We compare our model MPC to a variety of state-of-the-art approaches that perform multimodal fusion. We re-implement some methods to ensure they could perform flexible compositions of varying numbers of queries in arbitrary modalities, as described next.
\vspace{-0.5em}
\begin{itemize}[noitemsep,topsep=5pt,itemsep=2pt,leftmargin=10pt]
\item Relationship~\cite{santoro2017simple}: 
A relation reasoning module. 
It concatenates image features and text features, followed by an MLP to obtain a compositional embedding. 
\item FiLM~\cite{perez2018film}: 
A feature-wise linear modulation module. 
It learns affine transformation layers to modulate the image features based on the text features.  
\item MRN~\cite{kim2016multimodal}: 
A multimodal fusion module. 
It fuses the image and text features 
through three blocks of 
element-wise multiplication 
and residual learning. 
\item TIRG~\cite{Vo_2019_CVPR}: 
An multimodal composition module. 
It composes image and text features by concatenation, 
followed by a gating function 
and a residual connection for fusion. 
\item PCME + addition \cite{chun2021pcme}: It uses the same probabilistic embeddings in Eq. \ref{eq:prob_emb}. For composition, we add the two probabilistic embeddings directly, as defined below. 
\begin{equation}
\label{eq:pcme_addition}
\begin{aligned}
X &\sim \mathcal{N}(\mu_1, \Sigma_1); Y \sim \mathcal{N}(\mu_2, \Sigma_2)\\
X &+ Y \sim \mathcal{N}(\mu_1 + \mu_2, \Sigma_1 + \Sigma_2) = \mathcal{N}(\mu_c, \Sigma_c)\\
\end{aligned}
\end{equation}
\end{itemize}
\vspace{-0.5em}
\myparagraph{\underline{Discussion.}} 
Except for PCME, the other methods above were specifically designed to combine deterministic image and text features (either for VQA or for image search with text feedback). Our method is designed to leverage probabilistic embeddings and combine them in a way in which we can capture the polysemantic nature of the data. Our composition method is also agnostic of the input modality or the number of queries which allows us to combine any type of multimodal input. In our experiments, we reproduce the existing methods using the same ResNet backbone and pre-trained word embeddings for fair comparisons.

\myparagraph{Quantitative results.}
As mentioned in Section \ref{sec:setup}, we test various setups, including different input modalities, seen vs unseen compositions, and varying number of inputs. We present our experimental results on these setups below.

\myparagraph{Seen compositions with two inputs.} 
Table \ref{tab:2_categories_seen} shows our experimental results on composing two queries for multimodal image retrieval, where the model is tested on compositions seen during training. As can be seen, when given an arbitrary combination of the input modalities (i.e. images only, multimodal queries, texts only), we can observe that our model MPC outperforms the other competitors with substantial margins. On the compositions of image inputs, our MPC obtains a R@5 of 27.32\% vs 21.98\% by the best competitor TIRG. On the compositions of multimodal inputs, MPC obtains a R@5 of 37.76\% vs 24.59\% by the best competitor MRN. On the compositions of text inputs, MPC surpasses the best competitor TIRG by a margin of 24.53\% (59.32-34.79). 
Overall, our results show that MPC outperforms the other methods significantly, which indicates its strong generalization to digest an arbitrary combination of different input modalities for image retrieval. 

\begin{figure*}[!t]
\begin{center}
\includegraphics[width=.99\textwidth]{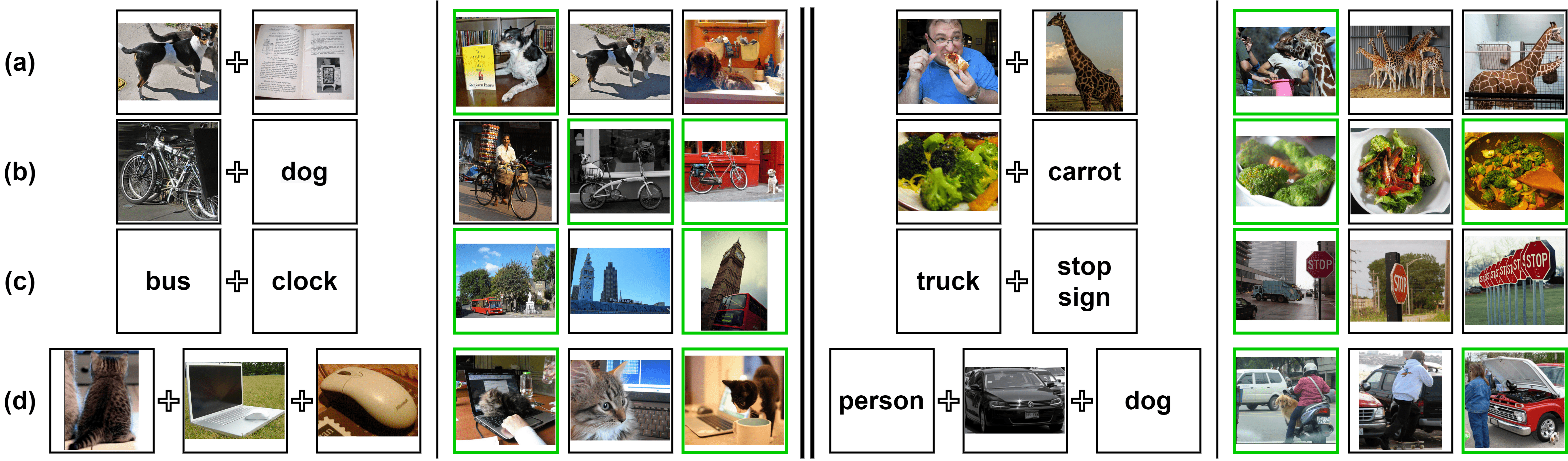}
\end{center}
\vskip -1.5em
\caption{Qualitative results of various types of multimodal queries: (a) image + image, (b) image + text, (c) text + text, (d) image + image + image / text + image + text.
For each example, we show the top 3 retrieved images and highlight the groundtruth with a \textcolor{OliveGreen}{green} box. 
}
\label{fig:qualitative}
\vskip -1.5em
\end{figure*}

\myparagraph{Seen compositions with three inputs.} 
Table \ref{tab:3_categories_seen} presents our experimental results on composing three queries for image retrieval, where we also test combinations of different input modalities. As shown, our MPC achieves the best performance on the compositions of multimodal and text-only inputs, obtaining R@5 of 19.26\%/32.52\% vs 17.63\%/25.16\% by the best competitor MRN. Although MPC does not surpass MRN on the compositions of image-only inputs, its performance is quite close to MRN, obtaining a R@5 of 8.90\% vs 9.89\% by MRN. From Table \ref{tab:2_categories_seen} and Table \ref{tab:3_categories_seen}, we observe that while the other methods could perform robustly well under different setups, MPC obtains the best overall model performance in varying number of inputs (2 or 3) and in different combinations of input modalities (image-only, text-only, or multimodal). Moreover, MPC surpasses others with great margins in handling text-only queries, thanks to the fact that it probabilistic nature can well capture the polysemantic semantics of words.

\begin{table}[!t]
	\small
	\centering
	\setlength{\tabcolsep}{10pt}
	\begin{tabular}{l|cc|c}
		\hline
		\multirow{2}{10pt}{Method} 
		& \multicolumn{3}{c}{\texttt{average}} \\ \cline{2-4}
		& R@5 & R@10 & R\textunderscore P  \\ 
		\hline
		 MRN & 
		 6.24 & 9.77 & 1.16 \\
		 TIRG & 
		 14.34 & 21.28 & 3.58 \\
		 PCME + addition & 
		 19.27 & 27.80 & 4.40 \\
		 \hline \hline
         \bf MPC  & 
		 \bf 26.11 &
		 \bf 37.41 &
		 \bf 6.05 \\
		\hline
	\end{tabular}
    \vskip -1em
	\caption{
	Evaluation of composing {\bf \em two} queries for image retrieval on a {\bf \em unseen} composition setup. 
	}
	\label{tab:2_categories_unseen}
	\vskip -1.5em
\end{table}

\myparagraph{Unseen compositions.} We further test a challenging setup with {\em unseen} compositions -- in similar spirit to compositional zero-shot learning \cite{purushwalkam2019task}, we consider the compositions of semantic concepts encountered at test time are unseen during training. Table \ref{tab:2_categories_unseen} shows the averaged results reported over all possible combinations of two input modalities (i.e. images only, texts only, multimodal queries), in comparison to three best competitors: MRN, TIRG, PCME + addition. We find that MPC outperforms other methods significantly, obtaining a R\textunderscore P of 6.05 vs 4.40 by the best competitor PCME + addition. Overall, these results suggest that MPC generalizes well at test time to digest unseen compositions of semantic concepts specified in queries.

\myparagraph{Qualitative results.} 
Figure \ref{fig:qualitative} shows the qualitative image retrieval results using a composite set of queries in different modalities. When given two inputs in arbitrary modalities (see (a), (b), (c)), our model can retrieve the images that contain the set of semantic concepts specified in the input, e.g. in the second example in (b), ``{\em broccoli}'' image and ``{\em carrot}'' text are composed to retrieve a dish with both concepts. When given three inputs (see (d)), our model can discover the images that cover the multiple semantic concepts specified in the input. Another interesting observation is that when given text queries, we find that our model is capable of capturing their polysemantic nature, e.g. "{\em clock}" may refer to a wall clock, but when composed with "bus", its semantic meaning becomes a clock tower, as shown in the 3 retrieved images in the first example in (c). 

\begin{figure*}[t]
\begin{center}
\includegraphics[width=.90\textwidth]{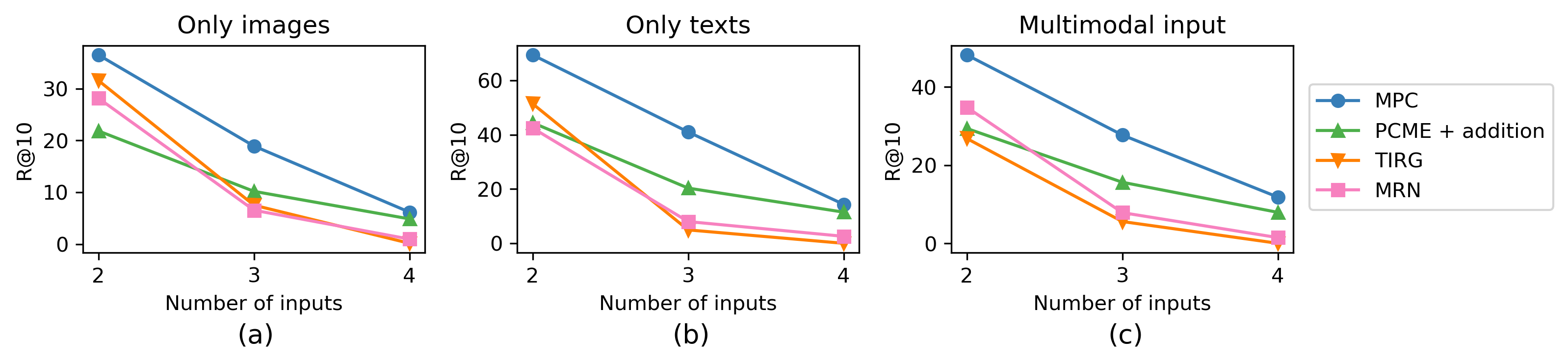}
\end{center}
\vskip -2em
\caption{
Evaluation of composing an increasing number of multimodal queries for image retrieval. All methods are trained to compose 2 inputs but tested to compose an increasing number of queries (i.e. 2, 3, 4). (a), (b), (c) show results on different combinations of input modalities. Note: as we increase the number of queries, the task difficulty is also increased. 
Numerical results are in the supplementary.
}
\label{fig:generalisation}
\vskip -1.em
\end{figure*}

\subsection{Model Analysis}
\label{sec:ablation}

\myparagraph{Generalization to more complex compositions.} As motivated, one uniqueness of our work is that we consider compositions upon an arbitrary number of queries. To test the generalization to handle an increasing number of inputs without additional training, we evaluate our model and three best competitors by training on compositions of 2 inputs and testing on compositions of 2, 3, 4 inputs. The more inputs are composed, the more challenging the task is, as the model needs to retrieve images that match all query concepts at once. Figure \ref{fig:generalisation} shows that under the varying number of inputs, our MPC maintains the best model performance under different combinations of input modalities. It is worth noting that the performance of non-probabilistic methods (MRN, TIRG) degrade to almost random guess ($\approx$0\% in R@5) when increasing the number of queries from 2 to 4. In contrast, the probabilistic methods (our MPC, PCME + addition) are able to retain a relatively reasonable model performance even when increasing the number of inputs to 3 or 4 without additional training. Overall, these results indicate that the probabilistic nature of our model helps to encode richer semantics and interactions between different semantic concepts, which strengthens its capability of composing an flexible amount of queries for image retrieval.

\begin{table}[!t]
	\small
	\centering
	\setlength{\tabcolsep}{10pt}
	\begin{tabular}{l|cc|c}
		\hline
		\multirow{2}{10pt}{Method} 
		& \multicolumn{3}{c}{\texttt{average}} \\ \cline{2-4}
		& R@5 & R@10 & R\textunderscore P  \\ 
		\hline
		 MPC w/o Eq. \ref{eq:gaussian_multiplication} & 
		 15.40 & 24.17 & 3.43 \\
		 MPC w/o Eq. \ref{eq:similarity_function} & 
		 30.26 & 38.62 & 8.05 \\
		 \hline
		 \bf MPC  & 
		 \bf 40.54 &
		 \bf 50.60 &
		 \bf 10.02 \\
		\hline
	\end{tabular}
    \vskip -1em
	\caption{
	Ablation study on our multimodal probabilistic composer.
	}
	\label{tab:ablations}
	\vskip -1.5em
\end{table}

\myparagraph{Ablation study.} 
To gain insights of our model, we analyze the following two components: our probabilistic composer (Eq. \ref{eq:gaussian_multiplication}) and our probabilistic similarity function (Eq. \ref{eq:similarity_function}). First, we compare our proposed probabilistic composer (Eq. \ref{eq:gaussian_multiplication}, Eq. \ref{eq:gaussian_multiplication2}) with a baseline ``MPC w/o Eq. \ref{eq:gaussian_multiplication}'' that fuses two embeddings by MLPs: $[\mu_c, \sigma_c^2] = f_{\text{MLP}}([\mu_1, \sigma_1^2, \mu_2, \sigma_2^2])$. 
Second, we compare our proposed probabilistic similarity function (Eq. \ref{eq:similarity_function}) with a baseline `MPC w/o Eq. \ref{eq:gaussian_multiplication}'' that uses an existing similarity function (Eq. \ref{eq:similarity_function_pcme}). 
Table \ref{tab:ablations} reports the averaged results over different combinations of input modalities. 
When comparing MPC and ``MPC w/o Eq. \ref{eq:gaussian_multiplication}'', we find that the model performance degrades significantly from 40.54\% to 15.40\% in R@5. This suggests that our probabilistic composer works much more effectively compared to simple composition with MLPs, owing to the rich inductive bias induced by our probabilistic composer. When comparing MPC and ``MPC w/o Eq. \ref{eq:similarity_function}'', we find that the model performance also decreases substantially from 40.54\% to 30.26\% in R@5. This indicates that our probabilistic similarity function offers a better estimated similarity as compared to the other similarity function, making our model converge well and learn a better embedding. 

\begin{figure}[!t]
\begin{center}
\includegraphics[width=0.7\linewidth]{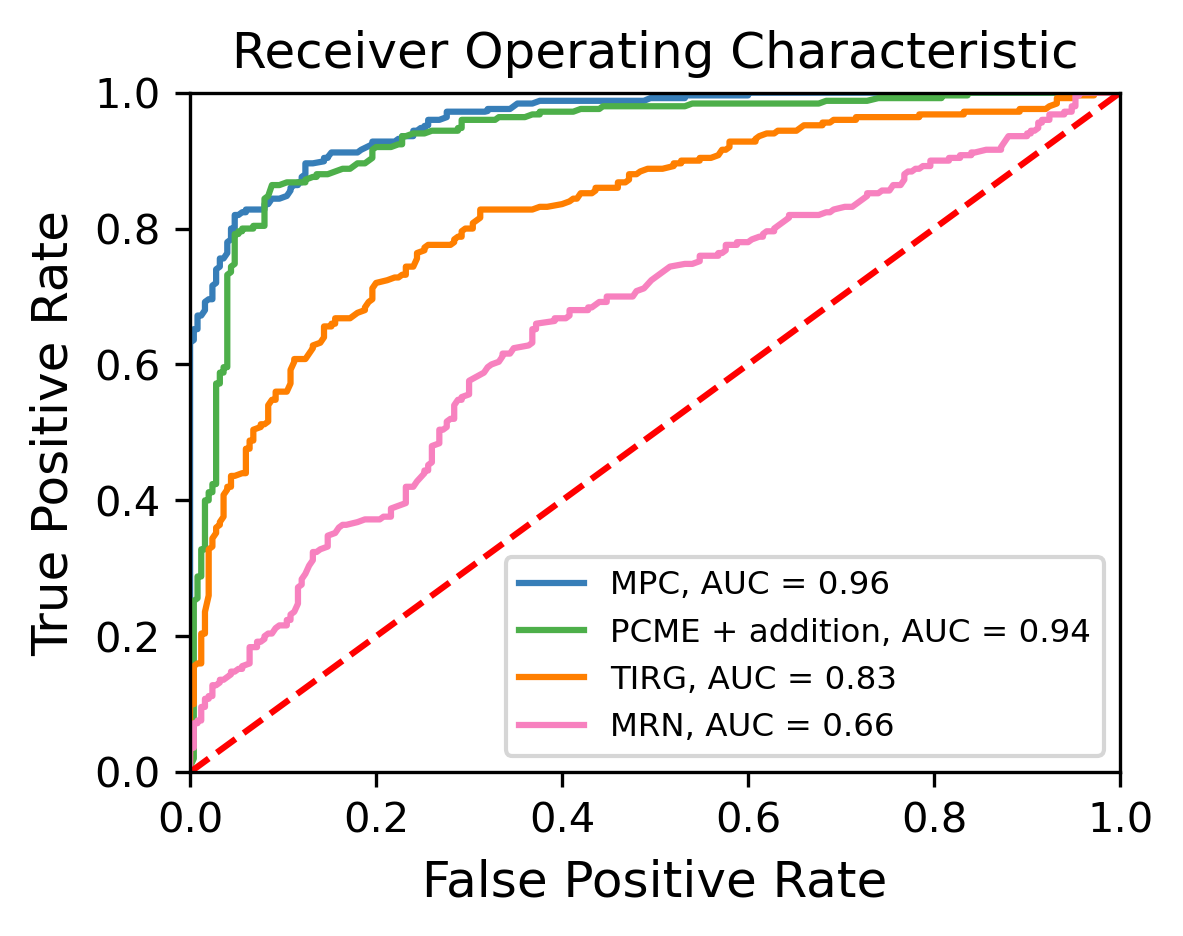}
\end{center}
\vskip -2em
\caption{The ROC curves and their AUC scores (area under curve, higher is better) for discriminating feasibility in compositions. 
}
\label{fig:feasibility}
\vskip -1.5em
\end{figure}

\myparagraph{Probabilistic uncertainty for discriminating feasibility.} 
The probabilistic nature of our model allows encoding both semantics and uncertainties for given inputs; while the {\em uncertainty} score of a probabilistic compositional embedding can be naturally used to indicate the {\em feasibility} in compositions -- 
it is feasible to compose ``{\em broccoli}'' and ``{\em carrot}'' as they can appear together in a dish; but it is infeasible to compose ``{\em broccoli}'' and ``{\em piano}''as they rarely appear together in real-world images. 
To utilize probabilistic uncertainty for discriminating feasibility, we use the Monte-Carlo estimation to derive uncertainty same as \cite{oh2018modeling}. 
For non-probabilistic models, we use the distance between two embeddings as the uncertainty score. We construct the feasible and infeasible sets of inputs based on their existence in MS-COCO dataset, where the feasible sets may be seen or unseen during training. 
We test the binary discrimination of feasibility based on ROC curve \cite{fawcett2006introduction}. Figure \ref{fig:feasibility} shows the comparison between MPC and the top competitors. 
We find that the probabilistic models (MPC, PCME + addition) obtain much higher AUC scores compared to the non-probabilistic models (TIRG, MRN); while our MPC obtains the best AUC of 0.96 among all methods. These results suggest that the uncertainty derived from MPC serves as a superior indicator for discriminating feasibility in compositions. 

\section{Conclusion and Future Work} 

In this work, we explored a new compositional learning paradigm for multimodal image retrieval. We established a new benchmark for multimodal image retrieval based on MS-COCO dataset, which poses a new challenge of composing a flexible amount of queries given in arbitrary visual and (or) textual modalities. To tackle this challenge, we proposed a novel multimodal probabilistic composer that learns probabilistic embeddings and composes embeddings based on a parametric probabilistic rule. We formulated a  probabilistic composer and a probabilistic similarity function to learn informative compositional embeddings. 
We demonstrated the superiority of our model compared to multiple existing methods under a diverse set of evaluation setups. 

Our model has shown its efficacy; however, there are still critical research problems to be solved for the new challenge posed in this work, such as considering more complex natural language queries, and ensuring model robustness when increasing the number of inputs. Moreover, to bring a positive societal impact, we also need to consider issues such as algorithmic biases and fairness when deploying our model to real-world search engines. 

\myparagraph{Acknowledgements} 
This work has been partially funded by the ERC (853489 - DEXIM) and by the DFG (2064/1 – Project number 390727645).

\setcounter{figure}{0}
\setcounter{section}{0}
\setcounter{table}{0}
\setcounter{algorithm}{0}
\renewcommand\thesection{\Alph{section}}
\renewcommand\thefigure{\Alph{figure}}
\renewcommand\thetable{\Alph{table}}
\renewcommand\thealgorithm{\Alph{algorithm}}

\vskip 1em
\noindent
{\Large \bf Supplementary materials}

\section{Additional Details in Benchmark Setup}

\myparagraph{Dataset and benchmark setup.} 
In order to construct our benchmark, we start from the 118,287 images in the training split of the MS-COCO dataset \cite{lin2015microsoft}. We split the images with a 4:1:1 ratio for the train/validation/test sets. 
Our benchmark is designed for the task of composing the concepts (which are object categories in this work) specified in the input queries for image retrieval. For this aim, we need to generate compositions of individual concepts that are well presented in a sufficient amount of images to serve for the retrieval. 
We choose 8:2:2 as the splitting thresholds for the train/validation/test splits. 
This means that a composition of concepts is valid only if it is present in at least 8 of the images in the training set, and 2 of the images in the validation and testing sets respectively. These values ensure that there is enough diversity in the training phase and also multiple correct answers (i.e. more than 1) in the evaluation phase. Using these thresholds allows us to find 1000 viable compositions for using either 2, 3 or 4 input queries. 
Algorithm \ref{alg:generate_combinations} details the process of generating a compositions of $k$ inputs. We will release our benchmark upon acceptance. 

\begin{algorithm*}[!t] 
\caption{Algorithm for generating a compositions of $k$ concepts (i.e. categories)}
\begin{algorithmic} 
\State compositions\_set $\gets$ empty set \Comment{initialize a set to store the compositions of $k$ concepts}
\State num\_of\_compositions $ \gets 0$ \Comment{record the number of found compositions in dataset}
\State target\_num\_of\_compositions $ \gets 1000$ \Comment{expected number of compositions to be found}
\State threshold\_\{train/val/test\} $\gets 8:2:2$   \Comment{minimal numbers of compositions in the train/val/test splits}

\While{num\_of\_compositions $< $  target\_num\_of\_compositions}
\State cur\_composition $\gets$ sample $k$ categories w/out replacement 
\Comment{get images for the current composition of $k$ concepts}
\State images\_\{train/val/test\} $\gets$ get \{train/val/test\} images that contain ``cur\_composition''
\If {len(images\_\{train/val/test) $\geq$ threshold\_\{train/val/test\}}
\If {cur\_composition not in compositions\_set} 
\State compositions\_set.insert(cur\_composition) \Comment{add a found composition to the set}
\State num\_of\_compositions $++$
\EndIf
\EndIf
\EndWhile
\end{algorithmic}
\label{alg:generate_combinations}
\end{algorithm*}

\myparagraph{Training.}
During training, for every concept (i.e. category) in the compositional inputs, we randomly choose an image id containing that concept, and also randomly pick either the visual or textual the modality to represent it. The target image is chosen randomly from the images that contain all the $k$ concepts specified in the inputs, but the modality is fixed to visual as we aim to solve an image retrieval task. 

\myparagraph{Evaluation.}
At test time, we evaluate for composing $k$ inputs (with $k$ fixed for an evaluation setup). The test queries are described by $k$ concepts and we generate all the test queries randomly to obtain a roughly equal mix of modality combinations. For example, for $k=2$, we will have around 25\% cases for each modality mixture in \{\textit{image-image}, \textit{image-text}, \textit{text-image},\textit{ text-text}\}. 

We also establish evaluation setups to test the model's ability of recognizing unseen compositions for retrieval, as well as testing the model's ability of identifying feasible/infeasible compositions. For the case of testing unseen compositions, we use $k=2$ and generate 100 compositions of different concepts for training and 500 new compositions of different concepts for testing, where the concepts at test time are seen during training but their compositions are unseen. We choose to use only 100 compositions for the training phase but 500 unseen compositions to simulate a more challenging evaluation setup. 
For testing the model's ability of identifying feasible/infeasible compositions, we build the evaluation setup upon the data generated for the seen compositions scenario for $k=2$. On top of the 1000 compositions, we generate an additional 250 unseen compositions and 250 infeasible compositions. An infeasible composition is defined as infeasible given that it is not found in any image across all the images in the dataset.

\begin{table*}[!t]
	\small
	\centering
	\setlength{\tabcolsep}{10pt}
	\begin{tabular}{l|ccc|ccc|ccc}
		\hline
        Input modalities
		& \multicolumn{3}{c|}{\texttt{images only}} 
		& \multicolumn{3}{c|}{\texttt{multimodal}} 
		& \multicolumn{3}{c}{\texttt{texts only}} \\ \cline{1-10}
	    Number of inputs & 2 & 3 & 4 & 2 & 3 & 4 & 2 & 3 & 4  \\
		\hline \hline
		 MRN~\cite{kim2016multimodal} & 
		 28.17 & 6.55 & 0.97 &
		 34.84 & 7.88 & 1.49 &
		 42.48 & 8.01 & 2.57 \\		 
		 TIRG~\cite{Vo_2019_CVPR} & 
		 31.58 & 7.52 & 0.16 &
		 26.85 & 5.60 & 0.03 &
		 51.43 & 4.94 & 0.00 \\
		 PCME~\cite{chun2021pcme} + addition & 
		 21.85 & 10.19 & 4.87 &
		 29.41 & 15.65 & 7.95 &
		 44.43 & 20.34 & 11.48\\
		 \hline \hline
        \bf MPC  & 
		 \bf 36.52 & 
		 \bf 18.93 & 
		 \bf 6.17 &
		 \bf 48.23 &
		 \bf 27.73 &
		 \bf 11.85 &
		 \bf 69.42 &
		 \bf 41.02 &
		 \bf 14.35 \\
		\hline
	\end{tabular}
	\vskip -1em
	\caption{
	Numerical results of model performance trained for composing 2 inputs but evaluated for composing 2, 3, and 4 inputs.  Results are accompanied to Figure 4 in the paper. Metrics: R@10 (\%). 
	}
	\label{tab:numerical_generalization}
	\vskip -1em
\end{table*}

\section{Numerical Results}
We present the numerical results used to generate Figure 4 in the main paper. Figure 4 shows the model performance when being trained to compose two inputs but tested on composing a varying number of inputs (i.e. 2, 3 and 4). We present these numerical results in Table \ref{tab:numerical_generalization}.

\begin{table*}[!ht]
	\small
	\centering
	\setlength{\tabcolsep}{10pt}
	\begin{tabular}{l|cc|c|cc|c|cc|c}
		\hline
		\multirow{2}{10pt}{Method} 
		& \multicolumn{3}{c|}{\texttt{images only}} 
		& \multicolumn{3}{c|}{\texttt{multimodal}} 
		& \multicolumn{3}{c}{\texttt{texts only}} \\ \cline{2-10}
		& R@5 & R@10 & R\textunderscore P & R@5 & R@10 & R\textunderscore P & R@5 & R@10 & R\textunderscore P  \\ 
		\hline
		 MRN~\cite{kim2016multimodal} & 
		 2.11 & 5.19 & 0.50 &
		 4.76 & 8.44 & 1.03 &
		 7.40 & 11.78 & 1.51 \\		 
		 TIRG & 
		 \bf 3.40 & \bf 6.98 & \bf 0.63 &
		 4.73 & 7.98 & 0.97 &
		 7.10 & 9.67 & 1.21 \\
		 PCME + addition & 
		 0.00 & 0.16 & 0.01 &
		 0.13 & 0.19 & 0.31 &
		 0.00 & 0.00 & 0.00 \\
		 \hline \hline
         \bf MPC  & 
		  2.76 & 
		  5.52 & 
		  0.59 &
		 \bf 5.98 &
		 \bf 10.03 &
		 \bf 1.30 &
		 \bf 8.46 &
		 \bf 16.77 &
		 \bf 1.95 \\
		\hline
	\end{tabular}
	\vskip -1em
	\caption{
	Evaluation of composing {\bf \em four} query inputs for image retrieval on a {\bf \em seen} composition setup. 
	}
	\label{tab:4_categories_seen}
	\vskip -1em
\end{table*}

\section{Additional Evaluation}
\myparagraph{Training with more inputs.}
In the main paper, we explored the scenarios where we train our model to compose two and three query inputs. However, our model formulation is not limited to a fixed number of inputs. Hence, we further explore the scenario where we train the models to compose four query inputs here. Table \ref{tab:4_categories_seen} contains the results for the top performing models when trained to compose four query inputs and be evaluated on a seen composition setup. As we can see, our model MPC achieves the best performance on the composition of multimodal and text-only inputs, obtaining a R@5 of 5.98\%/8.46\% vs 4.76\%/7.40\% by the best competitor MRN. Although MPC does not surpass TIRG on the compositions of image-only inputs, its performance is quite close to TIRG, obtaining a R@5 of 2.76\% vs 3.40\%. Another interesting observation is that while the other probabilistic model PCME+addition performs relatively well when composing two or three inputs, its performance degrades when composing four inputs. This suggests that composition with addition is weaker at preserving the information from more inputs, while our MPC model formulation with probabilistic composer can better capture the additive information from the increasingly more inputs.

\myparagraph{Evaluation on Fashion200k.}
In the main paper, we showed how our model is capable of composing queries of arbitrary sizes and modalities. In this section we want to showcase the performance of our model on the problem of image retrieval using image and attribute-based text feedback \cite{Vo_2019_CVPR, chen2020image, CoSMo2021_CVPR}. This is a slightly different setting than our original problem, as in this case the aim is to model the interactions between the image and text queries instead of adding them together. Fashion200k \cite{han2017automatic} is a dataset that contains around 200k images of fashion products. Each image is tagged with a set of attributes. Using these attributes, \cite{Vo_2019_CVPR} generated pairs of products that differ by only one attribute. The text modifications are generated using the attribute that is different between the 2 products. Table \ref{tab:fashion200k} contains the results on the Fashion200k dataset. We limited our comparisons to methods that are capable of modeling the original task in the paper. Other methods \cite{chen2020image, CoSMo2021_CVPR} have better performance, but use complex network architectures, specialized on this task, that can not be used to model arbitrary queries. Because of this, we chose to not add them to the comparison. As we can see, our model MPC achieves the best performance among the compared methods. We can also see that using simple additions of probabilistic embeddings (i.e. {PCME + addition}) is not capable of modeling the interactions between the image and text parts of the query, leading to poor results. These results show that our MPC also generalizes well on the other benchmark dataset Fashion200k.

\begin{table}[!t]
	\small
	\centering
	\setlength{\tabcolsep}{10pt}
	\begin{tabular}{l|ccc}
		\hline
		Method & R@1 & R@10 & R@50  \\ 
		\hline
		 Relationship & 
		 13.0 & 40.5 & 62.4 \\
		 Film & 
		 12.9 & 39.5 & 61.9 \\
		 MRN & 
		 12.3 & 39.4 & 60.9 \\
		 TIRG & 
		 14.1 & 42.5 & 63.8 \\
		 PCME + addition & 
		 1.8 & 11.4 & 27.3 \\
		 \hline \hline
         \bf MPC  & 
		 \bf 14.6 &
		 \bf 45.4 &
		 \bf 66.0 \\
		\hline
	\end{tabular}
	\vskip -1em
	\caption{
	Retrieval results on the Fashion200k dataset. 
	}
	\label{tab:fashion200k}
	\vskip -1.5em

\end{table}

{\small
\bibliographystyle{ieee_fullname}
\bibliography{reference}
}

\end{document}